\documentclass{svproc}
\usepackage{url}

\usepackage[numbers]{natbib}
\usepackage{colortbl}
\usepackage{xcolor}
\usepackage{booktabs}
\usepackage{graphicx}
\usepackage{subcaption} 
\usepackage{wrapfig}   

\usepackage{hyperref}
\usepackage{cleveref}
\usepackage{url}

\begin{document}
\mainmatter              
\title{Benchmarking Image Embeddings for E-Commerce: Evaluating Off-the Shelf Foundation Models, Fine-Tuning Strategies and Practical Trade-offs}
\titlerunning{Benchmarking Image Embeddings for E-Commerce}  
%

\author{
  Urszula Czerwinska\inst{1} \quad
  Cenk Bircanoglu\inst{1} \quad
  Jeremy Chamoux\inst{1} \\
}
\authorrunning{Czerwinska et al.} 
%
\tocauthor{Urszula Czerwinska, Cenk Bircanoglu, Jeremy Chamoux}
\institute{Adevinta, 22 rue des Jeûneurs, 75002 Paris, France\\
\email{{\{urszula.czerwinska, cenk.bircanoglu, jeremy.chamoux\}@adevinta.com}}}

\maketitle              

\begin{abstract}
We benchmark foundation models image embeddings for classification and retrieval in e-Commerce, evaluating their suitability for real-world applications. Our study spans embeddings from pre-trained convolutional and transformer models trained via supervised, self-supervised, and text-image contrastive learning. We assess full fine-tuning and transfer learning (top-tuning) on six diverse e-Commerce datasets: fashion, consumer goods, cars, food, and retail. Results show full fine-tuning consistently performs well, while text-image and self-supervised embeddings can match its performance with less training. While supervised embeddings remain stable across architectures, SSL and contrastive embeddings vary significantly, often benefiting from top-tuning. Top-tuning emerges as an efficient alternative to full fine-tuning, reducing computational costs. We also explore cross-tuning, noting its impact depends on dataset characteristics. Our findings offer practical guidelines for embedding selection and fine-tuning strategies, balancing efficiency and performance.
\end{abstract}

\begin{figure}[h]
\begin{center}
\includegraphics[width=0.8\textwidth]{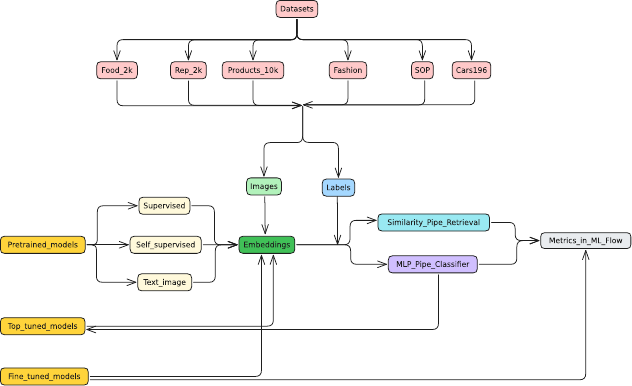}
\label{fig:general_schema}
\end{center}
\caption{\textbf{A high-level illustration of our experimental workflow.} We evaluate pre-trained, fine-tuned, and top-tuned models on six e-Commerce datasets, assessing performance through retrieval. Additionally, pre-trained models undergo classification testing via top-tuning. All metrics are logged in an MLflow dashboard.}
\end{figure}

\section{Introduction}
\label{sec:intro}

The rapid growth of e-Commerce has intensified the need for accurate and efficient image-based product categorization and retrieval. Machine learning (ML) and computer vision are essential for applications such as search optimization, recommendation systems, and automated product tagging \citep{Ye2022VISAtlas:}. While foundation models pre-trained image embeddings are widely used for their transferability, selecting the most effective embeddings and fine-tuning strategies for industry-specific applications remains an open challenge \citep{Xu2023Pretrained}.

Supervised learning has traditionally been the dominant approach, yet it can be computationally expensive and struggle with generalization across diverse product categories \citep{Villeneuve2023Framework}. In contrast, self-supervised learning (SSL) and contrastive learning offer scalable alternatives by learning meaningful representations without extensive labeled data \citep{caron2021emergingpropertiesselfsupervisedvision, ypsilantis2023universalimageembeddingslargescale, radford2021learningtransferablevisualmodels}. However, there is a lack of systematic evaluation comparing how different training modes, backbone architectures (convolutional vs. transformer-based), and fine-tuning strategies impact embedding performance in real-world e-Commerce applications.

To address this gap, we benchmark image embeddings from supervised, self-supervised, and contrastive pretraining foundation methods across six diverse open-source e-Commerce datasets. We evaluate both full fine-tuning and top-tuning, where additional layers are trained on frozen embeddings, to assess their effectiveness in classification and retrieval tasks \cref{fig:general_schema}.

Our results show that while full fine-tuning is consistently strong, contrastive text-image models can outperform it, and SSL embeddings can achieve competitive performance with lower computational cost. Additionally, top-tuning significantly enhances all model types while offering a cost-efficient alternative to full fine-tuning. We also analyze cross-tuning—applying top-tuning on a different dataset—highlighting its dataset-dependent effects.

This study provides practical insights for selecting foundation models embeddings and fine-tuning strategies in e-commerce applications, balancing computational efficiency with model performance.
\section{Related Work}  
\label{sec:related}  

Image embeddings are central to modern computer vision applications, including classification, retrieval, and multimodal tasks. We position our work within key developments in backbone architectures, self-supervised learning (SSL), universal embeddings, fine-tuning strategies, and domain-specific studies, highlighting the novelty of our contributions.  

{\bf Backbone Architectures and Representation Learning.} 
Backbone selection significantly impacts embedding quality. \citet{goldblum2023battlebackboneslargescalecomparison} compare convolutional and transformer-based architectures in various tasks but offer limited information on fine-tuning strategies for domain adaptation. In contrast, we systematically evaluate full fine-tuning, top-tuning, and cross-tuning in the context of e-Commerce.  

{\bf Self-Supervised Learning (SSL) Advances.}  
SSL reduces reliance on labeled data while achieving competitive performance \citep{he2021maskedautoencodersscalablevision, caron2021emergingpropertiesselfsupervisedvision, oquab2024dinov2learningrobustvisual}. However, SSL performance varies across tasks \citep{Lee2023Revisiting}. We extend this line of research by demonstrating how top-tuning enhances SSL embeddings, improving their adaptability in e-Commerce scenarios.  

{\bf Universal Image Embeddings.}
Universal embeddings, foundation model, such as ImageBind \citep{Girdhar2023ImageBind}, aim for broad applicability across vision, audio, and text. While these models offer versatility, their generalization to domain-specific tasks remains underexplored. Our study critically evaluates their performance in specialized e-Commerce applications.  

{\bf Supervised Learning as a Benchmark.} 
Supervised models like ResNet \citep{wightman2021resnetstrikesbackimproved} and ViT \citep{dosovitskiy2021imageworth16x16words} remain standard baselines but are computationally expensive. We assess supervised embeddings against SSL and contrastive learning approaches, highlighting cost-effective alternatives for e-Commerce tasks with imbalanced or sparse labels.  

{\bf Text-Image Embeddings and Multimodal Models.} 
Contrastive text-image models such as CLIP \citep{radford2021learningtransferablevisualmodels} enable robust multimodal understanding. Prior work by \citet{Rashtchian2023Substance} explores their semantic properties, but their effectiveness in pure image-to-image retrieval tasks is less examined. We provide a direct comparison of CLIP-based embeddings with traditional vision models, ensuring a fair evaluation by focusing on image retrieval rather than zero-shot or text-to-image tasks.  

{\bf Fine-Tuning Strategies for Domain Adaptation.}
Fine-tuning techniques, including top-tuning \citep{Alfano2022Fine-tuning, Tian2020Rethinking} and prompt-based methods \citep{arango2024quicktunequicklylearningpretrained}, offer efficient domain adaptation. FUNGI \citep{simoncini2024traingainselfsupervisedgradients} further explores label-free adaptation. We systematically evaluate fine-tuning and top-tuning, demonstrating the efficiency of top-tuning for e-commerce-specific classification and retrieval.  

{\bf Domain-Specific Embedding Studies.}
E-Commerce applications demand specialized embedding solutions \citep{liu2024transformerempoweredmultimodalitemembedding, marqoIntroducingMarqo} relaying on adaption of a foundational model. While prior work integrates text and image data for recommendations, we focus on systematically benchmarking different embedding paradigms for e-Commerce tasks, addressing dataset diversity and domain-specific nuances.  

{\bf Our Contributions.}  
Existing studies examine backbone architectures, pretraining paradigms, and fine-tuning, but few integrate these perspectives for real-world e-Commerce applications. Unlike prior work, which primarily benchmarks models on generic datasets, we provide a systematic evaluation of supervised, SSL, and contrastive learning embeddings in e-Commerce classification and retrieval tasks. Additionally, we analyze fine-tuning strategies, offering practical guidelines that balance performance and computational efficiency.  
\section{Methods}
\label{sec:methods}

\subsection{Preliminaries}

{\bf Image Embeddings in E-Commerce.}
Image embeddings encode visual data into compact vector representations, facilitating classification, retrieval, and recommendation tasks.

In our experiments, dataset images were preprocessed according to model requirements, with embeddings stored for downstream evaluation.

\subsection{ Machine Learning Tasks.}
We evaluate embeddings on two key tasks:

{\bf Classification.} Product images are categorized into predefined classes, enabling automated sorting and improved search. We assess classification using a small multi-layer perceptron (MLP) classifier trained on frozen embeddings. This classifier comprises 2-3 fully connected (FC) layers followed by a classification layer. Training consists of (1) Bayesian hyperparameter search over learning rate, momentum, number of FC layers, and optimizer settings (30 epochs, 2 repeats per trial), and (2) model training (1000 epochs with early stopping). We also evaluate cross-tuning, where embeddings from fully fine-tuned models are further adapted to new datasets. For classification, we use standard metrics such as accuracy, precision, recall, and f1-score.

{\bf Retrieval.} Given a query image, retrieval aims to find visually similar products. We index normalized embeddings into a vector database and use L2 distance for nearest-neighbor retrieval. Performance metrics (mMP@5, mR@1 as in \citet{ypsilantis2023universalimageembeddingslargescale}, MAP, MRR, NDCG) are computed, with $k=5$ unless otherwise defined. We evaluate pre-trained, fine-tuned, and top-tuned embeddings, assuming images from the same classification category are similar. This allows direct assessment of embeddings without a classification head.

\subsection{Fine-Tuning Techniques}
\label{sec:fine-tuning}

{\bf Full Fine-Tuning.}
All model parameters are updated during training on an e-Commerce dataset, allowing deep adaptation but increasing computational cost and overfitting risk. Models start from ImageNet pretraining, undergo standard preprocessing and augmentation, and are trained using a unified parameter set. Metrics from classification tasks are consolidated, and for retrieval, models are evaluated after removing classification heads.

{\bf Transfer Learning - Top-Tuning.}
A computationally efficient approach where pre-trained embeddings remain frozen while a small classifier (2-3 FC layers) is trained on top. This minimizes storage requirements, as embeddings can be stored in flat files, and enables quick adaptation without requiring access to original images.

{\bf Cross-Tuning.}
A discovery experiment testing whether fine-tuned embeddings generalize across datasets. A model is fully fine-tuned on dataset A, stripped of its classification layer, and then used to generate embeddings for dataset B, which are evaluated in a retrieval setup. This assesses whether domain adaptation from a related, larger dataset benefits retrieval on a different dataset.

\begin{table*}
\centering
\caption{Summary of datasets used in the experiments.}
\small
\begin{tabular}{lllrrrc}
\hline
\textbf{Dataset} & \textbf{Domain} & \textbf{Training Size} & \textbf{Test Size} & \textbf{Val Size} & \textbf{Categories} \\
\hline
Food2K & Food & 620,192 & 311,859 & 104,513 & 2000 \\
Cars196 & Cars & 8,144 & 8,041 & 8,041 & 196 \\
SOP & Online products & 59,551 & 60,502 & 60,502 & 12 \\
Rp2k & Retail products & 344,854 & 39,457 & 39,457 & 2384 \\
Product\_10k  & Retail products & 141,931 & 55,376 & 55,376 & 9691 \\
Fashion & Fashion and retail & 31,980 & 4,442 & 7,996 & 6 \\
\hline
\end{tabular}
\label{tab:datasets}

\end{table*}

\section{Datasets}

Our study utilizes multiple datasets spanning diverse product categories to ensure a robust evaluation of image embeddings in e-Commerce settings (\cref{tab:datasets}). A detailed description is provided in \cref{sec:app2}.

Fashion serves as a baseline dataset, acting as a technical control due to its relatively small size (~40k images) and limited number of classes (six). This makes it the simplest classification task. Product10k, Rep2k, and Food contain a large number of categories, introducing greater complexity. SOP and Cars represent medium-difficulty datasets, balancing dataset size and class diversity.

\begin{figure}[h]
\begin{center}
\includegraphics[width=0.9\linewidth]{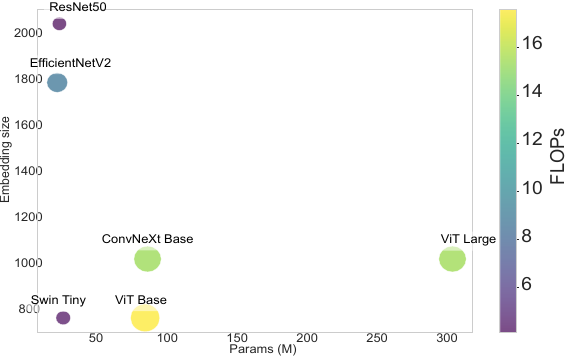}
\caption{Models used in this study, showing the relationship between embedding size, FLOPs (B), and parameters (M). DINO, DINOv2, MAWS, MAE, and CLIP share a ViT-B architecture and are represented alongside the vanilla ViT.}
\label{fig:modelsize}
\end{center}

\end{figure}

\section{Models}

To comprehensively assess the impact of backbone architecture, pre-training dataset, and training paradigm, we evaluate a diverse set of deep learning models \cref{fig:modelsize}. Our selection spans both supervised and self-supervised learning approaches, as well as contrastive text-image models.

\medskip\
{\bf Supervised Learning Models.} We include well-established architectures widely adopted in both research and industry. These consist of ViT-B, ViT-L \citep{dosovitskiy2021imageworth16x16words}, ConvNeXt-Base \citep{liu2022convnet2020s}, ResNet50 \citep{wightman2021resnetstrikesbackimproved}, EfficientNetV2 \citep{tan2020efficientnetrethinkingmodelscaling}, and Swin Transformer \citep{liu2021swintransformerhierarchicalvision}, all of which have demonstrated strong performance across a range of vision tasks.

\medskip\
{\bf Self-Supervised Learning Models.} We focus on state-of-the-art SSL models that have achieved competitive results in recent studies, including DINO \citep{caron2021emergingpropertiesselfsupervisedvision}, DINOv2 \citep{oquab2024dinov2learningrobustvisual}, MAE \citep{he2021maskedautoencodersscalablevision}, and MAWS \citep{singh2024effectivenessmaeprepretrainingbillionscale}. All SSL experiments are conducted using the ViT-B backbone, under the assumption that performance trends observed in ViT-B can be extrapolated to ViT-L.

\medskip\
{\bf Contrastive Text-Image Models.} We evaluate an extensive range of CLIP-style models that differ in pre-training datasets, parameter choices, and architectural variations. These include Meta CLIP \citep{metaclip_xu2024demystifyingclipdata}, EvaCLIP \citep{evaclip_sun2023evaclipimprovedtrainingtechniques}, Apple CLIP \citep{appleclip_vasu2024mobileclipfastimagetextmodels}, OpenAI CLIP (ResNet and ViT variants) \citep{radford2021learningtransferablevisualmodels}, SigLip \citep{siglip_zhai2023sigmoidlosslanguageimage}, and Marqo-B \citep{marqoIntroducingMarqo}, the latter specifically adapted for e-Commerce applications.

This diverse model selection allows us to systematically compare different architectural choices, pre-training strategies, and learning paradigms, offering valuable insights into their relative impact on classification and retrieval performance.

\section{Experiments and Results}
\label{sec:exp}

\subsection{Full Fine-Tuning}

For full fine-tuning, models were initialized with ImageNet-pretrained weights and trained following the selected procedure. Some models were subsequently saved without their classification head to generate embeddings, which were evaluated using the retrieval step of the benchmarking pipeline.

\medskip\
{\bf Classification Task.}
Following the A2 procedure \citep{wightman2021resnetstrikesbackimproved} with a batch size of 512 on four GPUs, results are presented in \cref{tab:results-fine-tun}. The best-performing model was ConvNeXt-Base, achieving 93\% accuracy, outperforming the second-best models (ViT-Base and DINO-ResNet50) by 3.6\%.
\begin{figure}[h]
\begin{center}

    \includegraphics[width=0.95\linewidth]{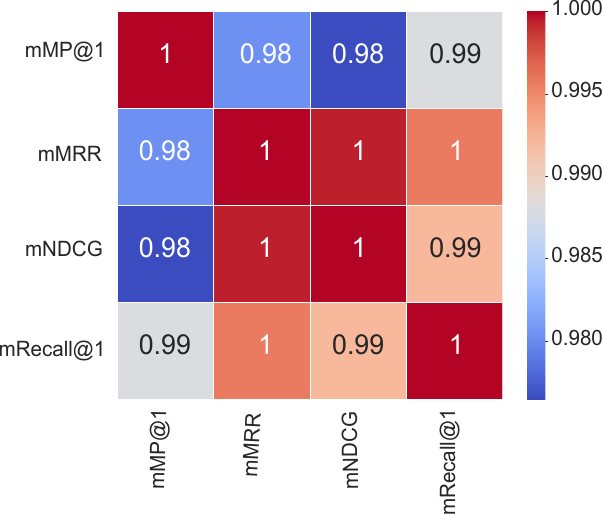}
    \caption{\textbf{Retrieval metric correlation.} All retrieval metrics used in this study are highly correlated.}
    \label{fig:metrics}
\end{center}
\end{figure}
Among self-supervised models, DINO-ResNet50 and MAE-ViT-B performed competitively, but did not surpass their supervised counterparts. ViT-L was the weakest supervised model, despite having the highest training cost (4.6× that of ViT-B). This suggests that the dataset size was insufficient for effective ViT-L training. Similarly, MAWS-ViT-B and DINO-ViT-B underperformed, indicating potential limitations in these SSL methods for this setup.

Classification accuracy closely correlates with retrieval performance (mMP@5), which we discuss in detail below \cref{fig:metrics}. Notably, results from the Cars196 dataset (\cref{tab:results-fine-tun}) reveal that self-supervised models exhibit significantly higher variance (0.08) compared to supervised models (0.0009, a 10× difference), suggesting greater instability in SSL embeddings for this dataset.

\begin{table}[h]
    \centering
    \small
    \caption{Performance metrics for different architectures on Cars196 dataset. Best values are bolded.}
    \renewcommand{\arraystretch}{1.2}
    \begin{tabular}{lrrr}
        \toprule
        Architecture & mR@1 (Label 5) & Val set accuracy & Training time (hrs) \\
        \midrule
        ConvNeXt-Base       & \textbf{0.924} & \textbf{0.930} & 2.67 \\
        DINO-ResNet50       & ---            & 0.898          & \textbf{1.75} \\
        ViT-Base-Patch16    & 0.886          & 0.898          & 2.42 \\
        MAE-ViT-Base        & ---            & 0.875          & 2.35 \\
        ResNet50            & 0.826          & 0.867          & 1.88 \\
        ViT-Large-Patch32   & 0.797          & 0.867          & 8.08 \\
        MAWS-ViT-Base       & ---            & 0.453          & 2.37 \\
        DINO-ViT-Base       & ---            & 0.344          & 2.45 \\
        \bottomrule
    \end{tabular}
    \label{tab:results-fine-tun}
\end{table}

{\bf Retrieval Task.}

\begin{figure}[t]
\small
    \centering
    \begin{subfigure}{\textwidth}
        \centering
        \includegraphics[width=\textwidth]{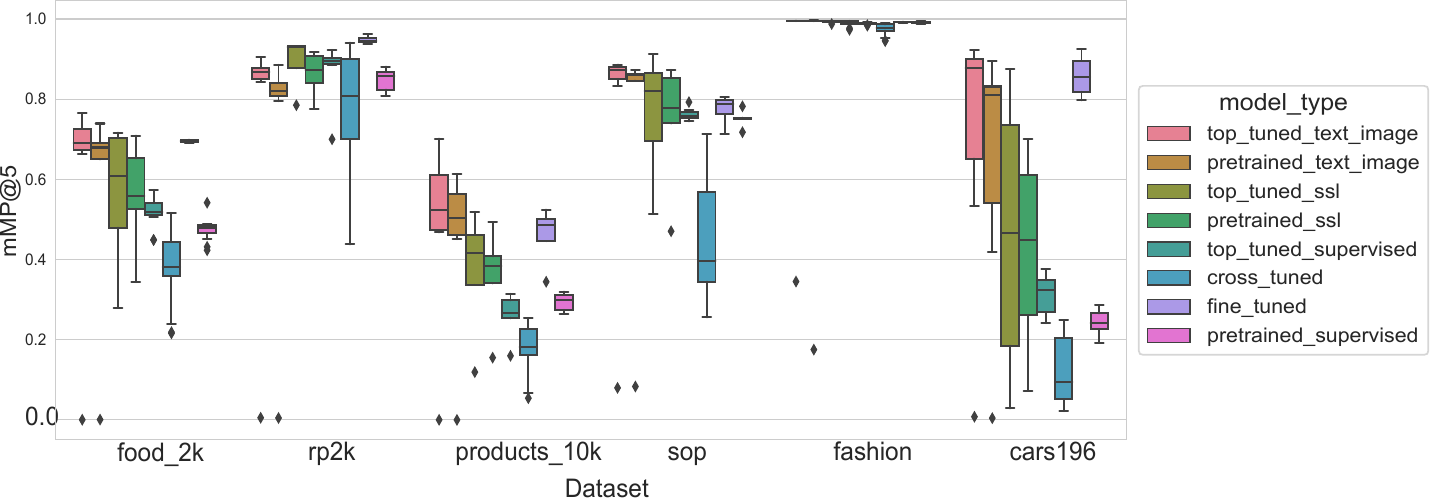}  
        \caption{Performance of all model types (mMP@5) on six datasets.}
        \label{fig:retrieval}
    \end{subfigure}
    \vspace{0.5em}

    \begin{subfigure}{0.44\textwidth}
        \centering
        \includegraphics[width=\textwidth]{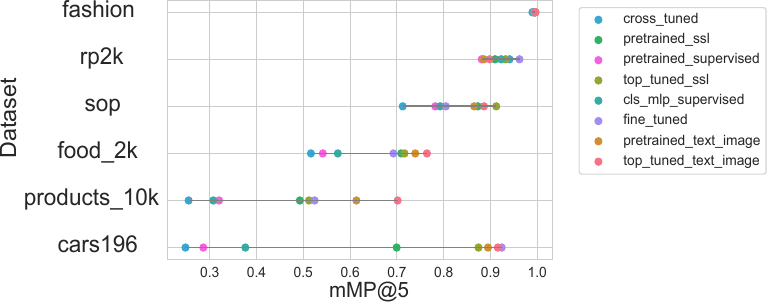}  
        \caption{Performance (mMP@5) of the best model of each type. Top-tuned text image are best in 4 datasets, fin-tuned in 1 dataset and top-tuned SSL in 1 dataset. The color legend same as in (a)}
        \label{fig:best-models}
    \end{subfigure}
    \hspace{2em} 
    \begin{subfigure}{0.44\textwidth}
        \centering
        \includegraphics[width=\textwidth]{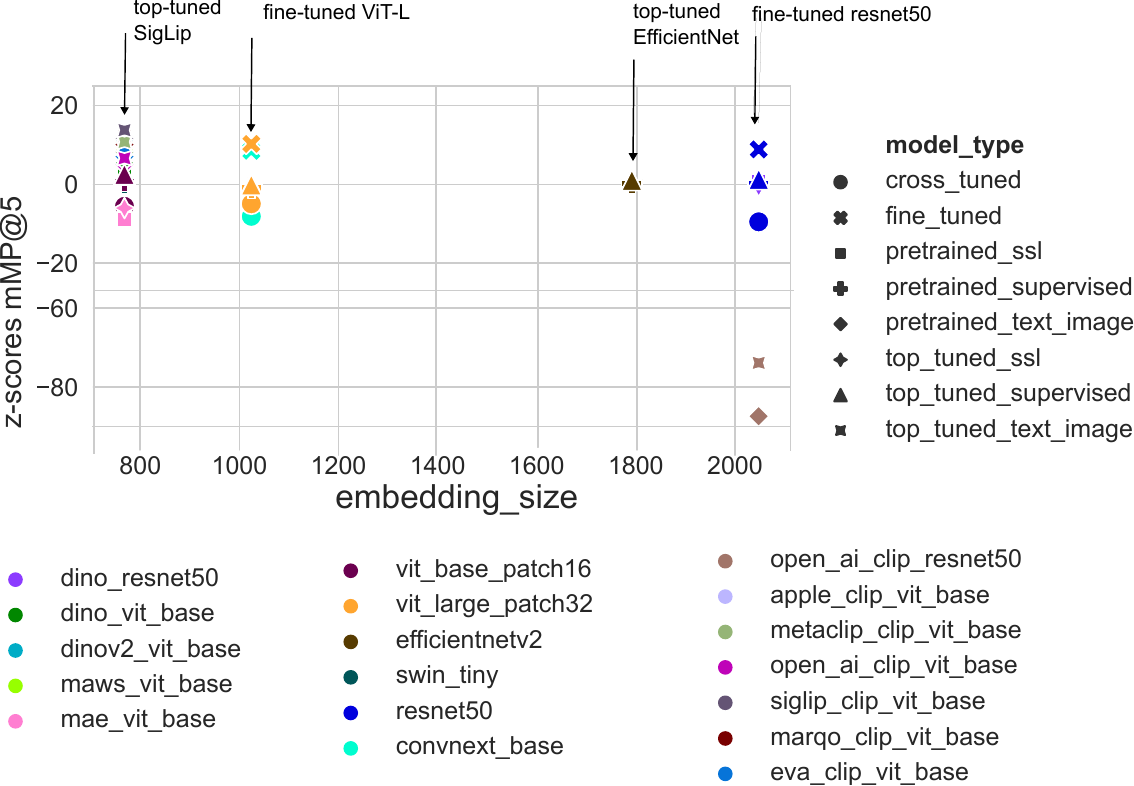}  
        \caption{Relative average performance of all architectures. Z-score was computed based on the mean performance for each dataset vs embedding size. Scale was adapted for better visual outcome.}
        \label{fig:size-perf}
    \end{subfigure}
    
      \caption{\textbf{All models results.} Retrieval performance comparison for all model types (mMP@5) (a), best models of each type hierarchy per dataset (b) and z-score normalized performance vs embedding size.}
    \label{fig:comparison}
\end{figure}

Fine-tuning on the target dataset consistently improves retrieval performance over pre-trained models (\cref{fig:retrieval}), often reaching or approaching state-of-the-art (SOTA) results previously reported in \citet{ypsilantis2023universalimageembeddingslargescale}. This highlights the necessity of dataset-specific fine-tuning for optimal retrieval performance.

Among fine-tuned supervised models (\cref{fig:retrieval}), retrieval performance varies between backbones. ConvNeXt-Base achieves the best results on three datasets (Cars196, SOP, Fashion) but performs the worst on Product-10k. ViT-B achieves top performance on two datasets and ranks second on two others, demonstrating strong generalization.

Training time is another key consideration. As shown in \cref{tab:results-fine-tun}, fine-tuning supervised models requires significantly more computational resources than self-supervised models. This variation is largely backbone-dependent, with ViT-L and ConvNeXt training considerably slower than ViT-B or ResNet50. Consequently, training time differences primarily stem from architectural choices rather than the pretraining paradigm.

\subsection{Pre-trained models}

In this section, we analyze the performance of different pre-trained embedding backbones and compare supervised and self-supervised pre-training approaches. Pre-trained embeddings are widely used due to their strong out-of-the-box performance, making them a convenient choice. Here, we evaluate their effectiveness on the retrieval task.

Supervised pre-trained models serve as a strong baseline for retrieval tasks \cref{fig:retrieval}. Performance variance across different backbones is relatively small. Notably, ViT-B achieved the best results in four out of six datasets \cref{fig:annex-all-bars}, while also offering an optimal balance between performance and embedding size. Surprisingly, ViT-L underperformed on the \texttt{product\_10k}, \texttt{cars196}, and \texttt{SOP} datasets. Additionally, while ConvNext is a strong model when fine-tuned, its out-of-the-box performance is weaker, ranking lowest among pre-trained models on \texttt{food2k} and second lowest on \texttt{product\_10k}.

As expected, the performance of supervised pre-trained models never surpasses that of fine-tuned ones. Only in the case of the relatively simple \texttt{Fashion} dataset does the pre-trained model match fine-tuned performance \cref{fig:retrieval}.

For self-supervised pre-trained models, we observe high performance variance across different backbones, similar to fine-tuned models \cref{fig:retrieval}. Among the SSL models, DINOv2 followed by MAWS achieves the best results, outperforming all other SSL models. Notably, there is little difference in performance between DINO with a ResNet-50 backbone and DINO with a ViT-B backbone, though the ViT-B variant consistently performs better. Compared to fine-tuned models, SSL pre-trained models achieve top performance depending on the dataset. Specifically, on \texttt{food2k}, \texttt{fashion}, \texttt{SOP}, and \texttt{product\_10k}, self-supervised pre-trained models outperform fine-tuned ones, with the most significant gains observed on \texttt{food2k} and \texttt{SOP}.

Finally, text-image pre-trained embeddings demonstrate strong performance on image-to-image retrieval benchmarks, though their performance varies across models \cref{fig:retrieval}. These models achieve the highest performance on \texttt{food2k}, \texttt{product\_10k}, and \texttt{cars196}, outperforming both supervised and self-supervised pre-trained models. Their performance on \texttt{rp2k}, \texttt{SOP}, and \texttt{fashion} is also competitive, making them a solid choice for out-of-the-box image retrieval tasks. Among text-image pre-trained models \cref{fig:annex-all-bars}, SigLIP performs best on five out of six datasets, with Apple CLIP outperforming it only on \texttt{rp2k}. Apple CLIP also emerges as a strong contender, ranking second-best among text-image models on three datasets. Additionally, on the \texttt{product\_10k} dataset, Marqo-B ranks as the second-best model and demonstrates strong performance across \texttt{SOP}, \texttt{cars196}, \texttt{fashion}, \texttt{food2k}, and \texttt{product\_10k}.

\subsection{Top-tuned models}

We define top-tuning as a transfer learning approach where two or three fully connected linear layers, including a classification layer, are added on top of pre-trained embeddings extracted from the penultimate layer of a pre-trained model. This method is both time- and cost-efficient, making it an accessible solution for data science teams.

\begin{table}[t]
    \small
   
    \label{tab:performance_top_tuned}
    \centering
    \caption{Performance comparison of different models with improvements shown in green.}
    \begin{tabular}{lrrrrrrr c}
        \toprule
        \textbf{Model Type} & \textbf{Cars196} & \textbf{Fashion} & \textbf{Food\_2k} & \textbf{Prod\_10k} & \textbf{Rp2k} & \textbf{SOP} & \textbf{Mean} & \textbf{} \\
        \midrule
        Pretrained Supervised  & 0.249 & \textbf{0.991} & 0.470 & \textbf{0.310} & 0.855 & 0.762 & 0.606 &  \\
        \rowcolor{gray!15} Top-tuned Supervised  & \textbf{0.355} & \textbf{0.991} & \textbf{0.544} & 0.298 & \textbf{0.908} & \textbf{0.775} & \textbf{0.645} & \textcolor{green}{↑3.9} \\
        Pretrained SSL  & 0.533 & \textbf{0.994} & 0.640 & 0.428 & 0.898 & 0.834 & 0.721 &  \\
        \rowcolor{gray!15} Top-tuned SSL  & \textbf{0.692} & \textbf{0.994} & \textbf{0.676} & \textbf{0.465} & \textbf{0.931} & \textbf{0.866} & \textbf{0.771} & \textcolor{green}{↑5.0} \\
        Pretrained Text-image  & 0.847 & 0.995 & 0.695 & 0.581 & 0.839 & 0.860 & 0.803 &  \\
        \rowcolor{gray!15} Top-tuned Text-image  & \textbf{0.907} & \textbf{0.996} & \textbf{0.739} & \textbf{0.641} & \textbf{0.887} & \textbf{0.882} & \textbf{0.842} & \textcolor{green}{↑3.9} \\
        \bottomrule
        
    \end{tabular}
\end{table}

Top-tuning proves to be an effective technique across the datasets we tested, particularly when applied to self-supervised model embeddings \cref{tab:performance_top_tuned}. In the case of supervised pre-trained models, top-tuning improves performance on all datasets except \texttt{product\_10k}, leading to an average improvement of 3.9\%. However, the results still fall short of fully fine-tuned models \cref{fig:retrieval}.

For text-image models, top-tuning consistently improves performance across all datasets \cref{tab:performance_top_tuned} and \cref{fig:annex-delta-all}. Since these models already perform well out of the box, the relative improvement is smaller. However, top-tuning allows text-image models to match or surpass supervised fine-tuning on four out of six datasets.

The most significant performance gains are observed when applying top-tuning to self-supervised models (mean 5\%) \cref{tab:performance_top_tuned} and \cref{fig:annex-heat-pretrained-ssl}, \cref{fig:annex-pretrained-ssl}.The overall improvement over pre-trained self-supervised models is the highest among all model types, with performance reaching or even exceeding fine-tuned models in some cases—such as on \texttt{SOP}. However, the effectiveness of top-tuning varies depending on the model architecture. Specifically, it yields positive results for \texttt{dino\_vit}, \texttt{dinov2\_vit}, and \texttt{maws} on most datasets (except for \texttt{maws} on \texttt{SOP}). Conversely, top-tuning negatively impacts performance for \texttt{dino\_resnet\_50} (-6.67\% in average) and \texttt{mae} (-15.09\% in average) across most datasets \cref{fig:annex-heat-pretrained-ssl}.

The largest improvement is observed on \texttt{cars196}, the smallest and most specialized dataset. This suggests that top-tuning can help pre-trained models specialize for specific tasks. However, its impact should always be compared against the baseline performance of the pre-trained model to determine whether it is a beneficial adaptation.

\subsection{Cross-top-tuned models}

Fine-tuning on one dataset and applying the model for retrieval on a different dataset typically leads to a significant decline in performance \cref{fig:annex-delta-cross} (up to -0.5 mMP@5). However, when the datasets share similar characteristics, cross-top-tuning can have a positive effect (up to ~0.1 mMP@5). For example, models fine-tuned on \texttt{products\_10k} or \texttt{cars196} achieve strong results on \texttt{rp2k}. Similarly, for \texttt{food2k}, a model fine-tuned on \texttt{products\_10k} attains performance comparable to the best pre-trained model.

\section{Comparison to prior work}  
\label{sec:comparison}  

The benchmarking of image embeddings for classification and retrieval has been extensively studied, with prior work comparing pre-training paradigms across diverse domains. Our study builds on this foundation by systematically evaluating fine-tuning strategies, self-supervised learning (SSL), and contrastive text-image models in an e-commerce context, an area with distinct challenges such as high inter-class similarity, long-tailed distributions, and fine-grained retrieval needs.  

\subsection{Embedding benchmarking and performance}  

The large-scale analysis by~\citet{goldblum2023battlebackboneslargescalecomparison} highlights the strengths of convolutional and transformer-based backbones under different pre-training paradigms. Consistent with their findings, we observe that Vision Transformers (ViTs) fine-tuned on domain-specific data consistently outperform convolutional models. However, our results go further by demonstrating that top-tuning SSL and text-image embeddings can yield performance better than or same as full fine-tuning while reducing computational costs.  

Unlike prior benchmarks, we show that contrastive text-image embeddings, such as SigLIP, achieve state-of-the-art performance on several retrieval tasks without requiring domain-specific fine-tuning. This suggests that contrastive multimodal pretraining provides robust visual representations even for pure image retrieval, a finding that prior studies have not systematically explored.  

\subsection{Self-Supervised Learning}  

SSL has gained prominence by reducing dependence on labeled data while achieving competitive results across various tasks \citep{caron2021emergingpropertiesselfsupervisedvision, he2021maskedautoencodersscalablevision}. However, prior work has noted the high variability in SSL performance across domains \citep{Lee2023Revisiting}. Our results reinforce this, showing that SSL embeddings exhibit higher variance than supervised models, particularly in retrieval tasks.  

Crucially, we find that top-tuning significantly stabilizes SSL embeddings, particularly for DINOv2 and MAWS, narrowing the performance gap with fully fine-tuned supervised models. This supports the argument that SSL models should not be dismissed due to their raw out-of-the-box variability, as lightweight adaptation strategies can enhance their effectiveness with minimal computational overhead.  

\subsection{Fine-Tuning Strategies}  

Fine-tuning is well-established for domain adaptation, with full fine-tuning often assumed to be necessary for optimal performance \citep{wightman2021resnetstrikesbackimproved, liu2024transformerempoweredmultimodalitemembedding}. However, its high computational cost limits its scalability. Recent studies explore alternatives such as top-tuning~\citep{Alfano2022Fine-tuning} and prompt-based tuning~\citep{arango2024quicktunequicklylearningpretrained}, but systematic evaluations in e-commerce contexts remain sparse.  

Our results provide a direct comparison of fine-tuning strategies, revealing key trade-offs:  
(1) \textbf{Full fine-tuning} remains the strongest approach but is computationally expensive. (2) \textbf{Top-tuning} significantly improves SSL and text-image embeddings, often matching full fine-tuning in retrieval tasks while requiring far fewer resources. (3) \textbf{Cross-tuning} shows mixed effectiveness, with gains dependent on dataset similarity, limiting its generalizability.  

This suggests that lightweight adaptation strategies are particularly effective for contrastive embeddings, an insight not covered in previous work.  

\subsection{E-Commerce Context}  

Most prior studies evaluate embeddings on standard datasets such as ImageNet and COCO~\citep{goldblum2023battlebackboneslargescalecomparison}, which do not fully capture the complexities of e-commerce. Our study fills this gap by benchmarking embeddings across six diverse e-commerce datasets, covering domains such as fashion, retail, food, and automobiles.  

Our results reveal that general-purpose pre-trained models do not always perform optimally in e-Commerce settings, particularly in fine-grained retrieval. In contrast, contrastive text-image models—previously optimized for multimodal tasks—demonstrate strong performance in pure image-to-image retrieval, challenging assumptions about their domain specificity. 
Surprisingly, Marqo-B \citep{marqoIntroducingMarqo} that was shown to beat its baseline SigLip on zero shot text to image retrieval (labels and categories) does not beat SigLip on image to image retrieval on our benchmark.

Overall, our findings refine prior understanding of embedding selection and adaptation, emphasizing that text-image models and SSL embeddings can achieve state-of-the-art performance in e-Commerce with minimal fine-tuning, significantly reducing computational costs while maintaining retrieval effectiveness.

\section{Conclusion}

Compared to previous studies, our work makes the following key contributions:
(1) \textbf{Fine-Tuning Strategies in Practice}: We go beyond standard embedding benchmarking by systematically evaluating full fine-tuning, top-tuning, and cross-tuning. Our results provide actionable guidelines for balancing accuracy and computational efficiency in real-world deployments.
(2) \textbf{Contrastive Text-Image Models for Image Retrieval}: Unlike prior work focused on zero-shot learning, we demonstrate that contrastive models (e.g., SigLIP, Marqo-B) achieve state-of-the-art performance in pure image-to-image retrieval, often outperforming fully fine-tuned supervised models.
(3) \textbf{Cross-Tuning Analysis}: While most studies focus on direct fine-tuning, we evaluate cross-tuning as a transfer strategy and highlight its dataset-dependent limitations, offering a more nuanced understanding of when adaptation across domains is effective.

Dataset size and granularity strongly influence adaptation strategies, with smaller datasets benefiting the most from top-tuning. Notably, the underperformance of MAE embeddings as frozen features suggests they encode more raw visual information, requiring extensive adaptation for semantic tasks. Despite these insights, limitations remain—our results may not fully generalize beyond e-Commerce, and computational constraints restricted large-scale evaluations. Future research should explore hybrid fine-tuning strategies, automated embedding selection frameworks, and broader multimodal applications in industry. For the practical aspect of usability of embeddings in industrial application, one could investigate the effect of distillation and quantization on embeddings performance.

Overall, our contributions refine the current understanding of embedding selection and adaptation, providing practical guidance for deploying vision models in real-world e-Commerce systems.

\subsubsection*{Acknowledgments}
We thank \textbf{X} for their valuable feedback and insightful suggestions that significantly improved the clarity and quality of this work.  

Additionally, we express our gratitude to \textbf{Company Name} for providing computational resources and support throughout this research.

%
%

\bibliography{template}
\bibliographystyle{bibtex/spbasic}

\appendix
\section{Appendix}

\subsection{Appendix I: Supplementary figures}
\label{sec:app1}

\begin{figure}[htbp]
  \centering
  \includegraphics[width=0.8\textwidth]{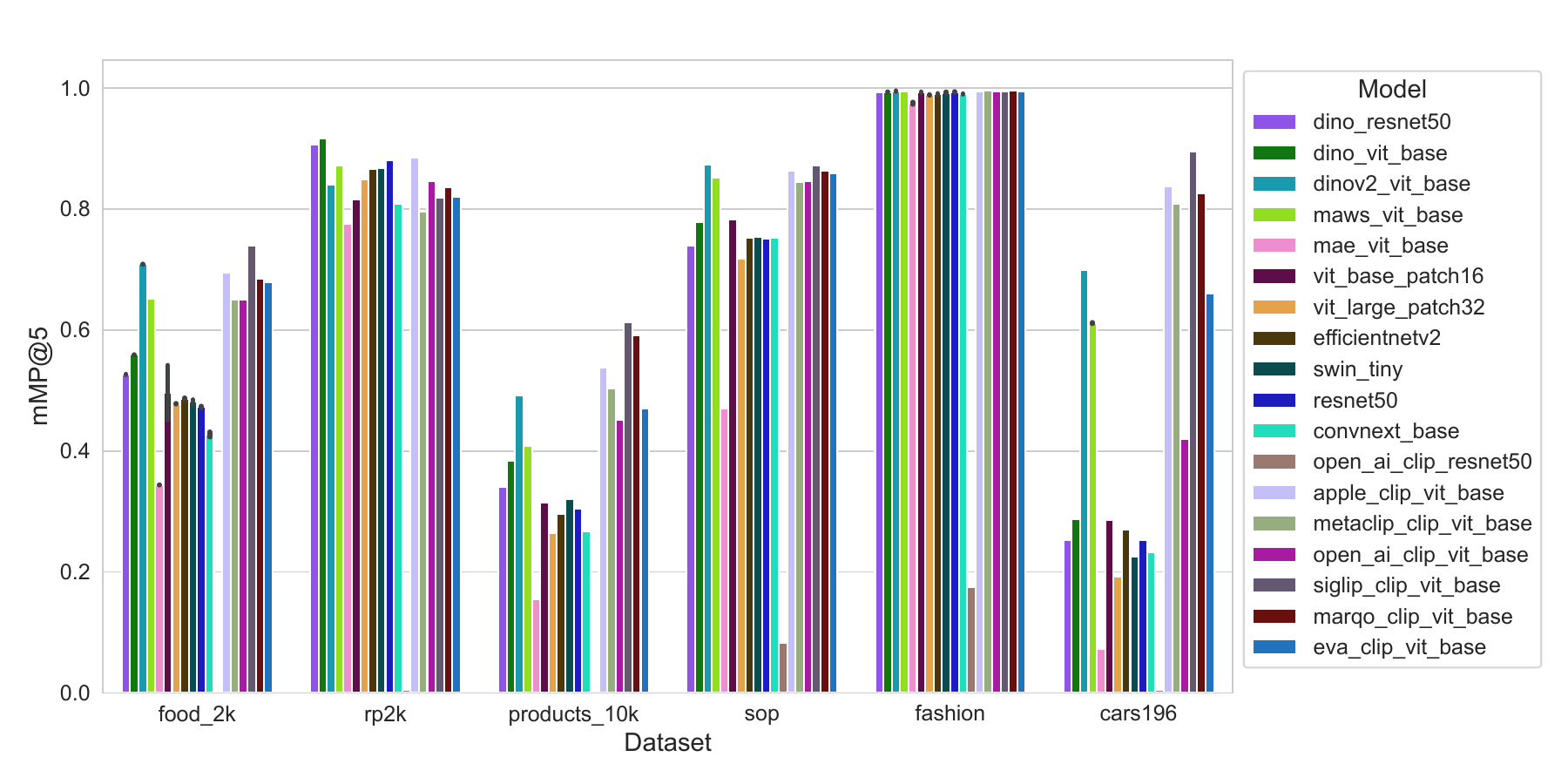} 
  \caption{Detailed results of all pretrained models performance on each dataset.}
  \label{fig:annex-all-bars}
\end{figure}

\begin{figure}[htbp]
\label{fig:annex-delta-all}
  \centering
  \begin{subfigure}[b]{0.45\textwidth}
    \centering
    \includegraphics[width=\linewidth]{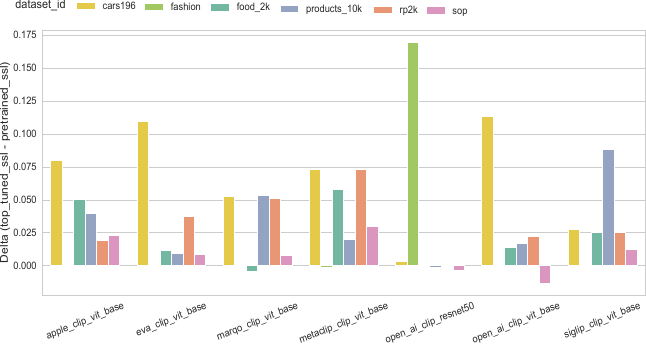} 
    \caption{Text-image models. The delta between pre-trained model performance and top-tuned model performance of each model for each dataset}
    \label{fig:annex-pretrained-text-im}
  \end{subfigure}
  \hfill
  \begin{subfigure}[b]{0.45\textwidth}
    \centering
    \includegraphics[width=\linewidth]{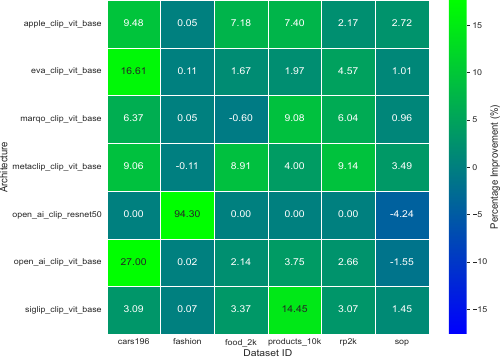} 
    \caption{Text-image models. The percentage of gain/loss for the pre-trained model after top-tuning.}
    \label{fig:annex-heat-pretrained-text-im}
  \end{subfigure}
  
  \vspace{1em} 
  
  \begin{subfigure}[b]{0.45\textwidth}
    \centering
    \includegraphics[width=\linewidth]{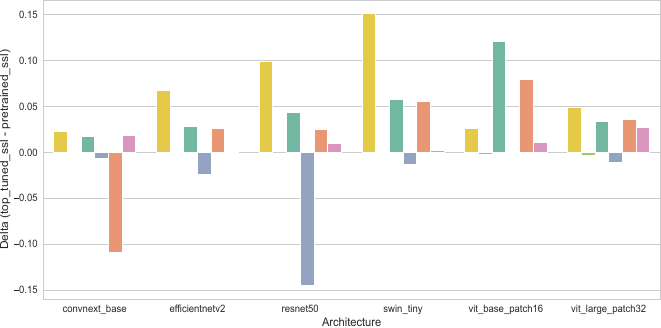} 
    \caption{Supervised models. The delta between pre-trained model performance and top-tuned model performance of each model for each dataset}
    \label{fig:annex-pretrained-sup}
  \end{subfigure}
  \hfill
  \begin{subfigure}[b]{0.45\textwidth}
    \centering
    \includegraphics[width=\linewidth]{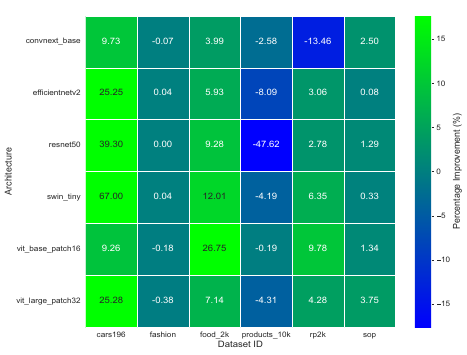} 
    \caption{Supervised models. The percentage of gain/loss for the pre-trained model after top-tuning.}
    \label{fig:annex-heat-pretrained-sup}
  \end{subfigure}
  
  \vspace{1em}
  
  \begin{subfigure}[b]{0.45\textwidth}
    \centering
    \includegraphics[width=\linewidth]{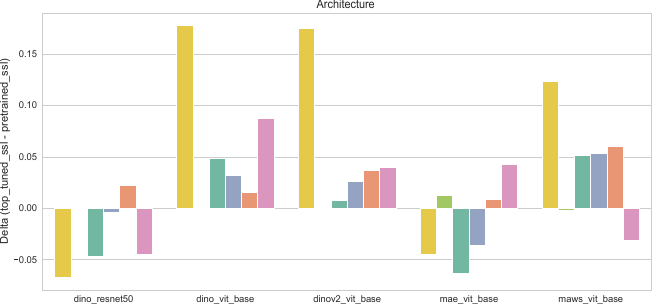} 
    \caption{Self-supervised models. The delta between pre-trained model performance and top-tuned model performance of each model for each dataset}
    \label{fig:annex-pretrained-ssl}
  \end{subfigure}
  \hfill
  \begin{subfigure}[b]{0.45\textwidth}
    \centering
    \includegraphics[width=\linewidth]{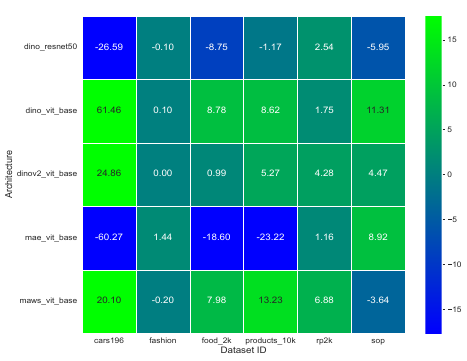} 
    \caption{Self-supervised models. The percentage of gain/loss for the pre-trained model after top-tuning.}
    \label{fig:annex-heat-pretrained-ssl}
  \end{subfigure}
  
  \caption{The detailed analysis of the top-tuning impact on the pretrained models.}
  \label{fig:overall}
\end{figure}

\begin{figure}[htbp]
  \centering
  \includegraphics[width=0.8\textwidth]{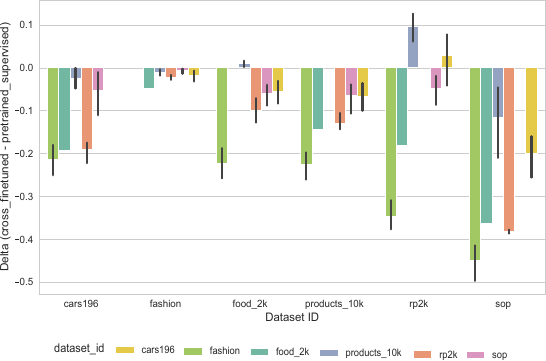} 
  \caption{Cross-tuning. The difference between the cross-tuned and pretrained model in mMP@5 among all tested backbones. X-axes: the top-tuning dataset, color bars correspond to the fine-tuning dataset.}
  \label{fig:annex-delta-cross}
\end{figure}

\clearpage

\subsection{Appendix II: Datasets}
\label{sec:app2}
\textbf{Stanford CARS196}: A dataset of 16,185 images across 196 car categories, used for fine-grained classification and visual recognition. Each category corresponds to a distinct car model, with images that vary in viewpoint, lighting, and background.

\textbf{Stanford Online Products (SOP)}: Contains 120,000+ images of products from 12 categories, designed for metric learning tasks such as product retrieval and visual search. Each product is represented by multiple images from different angles.

\textbf{Rp2k}: A large-scale dataset with 2,000 object categories, used for object detection and segmentation tasks. It contains thousands of images per category, supporting research in 3D object recognition.

\textbf{Product-10k}: A dataset with 10,000 product categories and images sourced from e-Commerce platforms. It is used for large-scale product recognition and retrieval tasks in real-world conditions.

\textbf{Fashion Product Images Dataset}: A dataset containing over 44,000 high-resolution images of fashion products from six categories (e.g., tops, pants, shoes). It includes product images along with additional metadata such as brand, price, and product description, designed for tasks like product classification and retrieval in e-Commerce applications.

\subsubsection{Dataset-Specific Insights}
The evaluation across e-Commerce datasets underscores the importance of domain-specific considerations. For instance:
\begin{itemize}
    \item Smaller datasets with high inter-class separability benefit most from top-tuning.
    \item Highly granular datasets, such as Product-10k, often require full fine-tuning for optimal performance.
\end{itemize}
These observations suggest that embedding selection and tuning must be tailored to the characteristics of the target dataset.

\subsection{Appendix III: Implementation details}
{\bf Image preprocessing}
The images were pre-processed following standard procedures for each model type. Typically provided together with timm library. For models from facebookresearch (torchhub) we applied standard scaling and normalisation. We have tested different mean and variance values for the mae model (given its low performance) but we did not observed significant difference.

{\bf Tuner configuration}
For the tuner we used `kerastuner` library. We used bayesian search for following hyperparameters: learning\_rate, number of hidden layers, number of units in hidden layers, activation function, optimiser and clip value. The Tuner runs for 30 epochs for each trial, given 20 trials, each redone twice. As loss function Cross Entropy loss is used. The best model is saved and the training continues from saved checkpoint for the final model keeping best parameter set.

{\bf Full fine-tuning configuration}
Selecting the right training parameters is crucial for optimizing model performance. To determine the best configuration, we conducted a series of experiments inspired by \citet{wightman2021resnetstrikesbackimproved}, identifying the most suitable training procedure for our setup. Given budget and time constraints, we focused on the three top-performing strategies from \citet{wightman2021resnetstrikesbackimproved}: A1, A2, and A3.

In our experiments, we tested these three procedures while making two modifications: replacing Binary Cross-Entropy (BCE) with Cross-Entropy (CE) and omitting Stochastic Depth. The decision to use CE was based on the findings in \citet{wightman2021resnetstrikesbackimproved}, which reported no significant difference between the two loss functions. As for Stochastic Depth, we excluded it to maintain compatibility with pre-trained models, which we intended to use.

Beyond comparing training procedures, we observed that batch size and the number of GPUs significantly impacted performance—particularly for the Cars196 dataset. A batch size of 512, combined with 8 GPUs, produced the best results. Additionally, training a ResNet50 model from scratch yielded noticeably worse performance compared to initializing with pre-trained weights. We attribute this to the limited size of the Cars196 training split, which contains only ~8,000 images.

Ultimately, the best results were obtained using the A1 procedure with a batch size of 512 and 4 GPUs. However, the training process took longer than expected, requiring 600 epochs. To balance efficiency and accuracy, we selected A2 with a batch size of 512 and 4 GPUs as our final approach. This configuration resulted in only a 1\% drop in validation accuracy compared to the best-performing model while significantly reducing training time.

{\bf Hardware Specifications}

\textbf{CPUs and GPUs}: The training was performed on Sagemaker, using 4 Nvidia A10G GPUs. Other experiments were conducted on Kubeflow pipelines using a minimum of 2 CPUs with 8 GB RAM. When necessary, computations were accelerated with 2 GPUs of type g4dn.xlarge, each with 12 GB memory.

\textbf{Cluster Configuration}: Kubeflow pipelines were managed with the "unicorn kfp-unicron" setup, version 2.0.0.

\textbf{Software and Libraries}
\begin{itemize}
    \item \textbf{Operating System}: Unix-based systems were used for all experiments.
    \item \textbf{Data Pre-processing}: Pre-processing tasks were performed using PyTorch (\texttt{torch==2.1.2}).
    \item \textbf{Model Implementation}: Models were sourced from the \texttt{timm} library (\texttt{timm==1.0.7}) or \texttt{torchhub} (\texttt{torchhub==xxx}) and registered in \texttt{mlflow} (\texttt{mlflow==2.3.0}) registry. All of the models were configured to return embeddings and classification head was removed. Embeddings were generated in Pytorch (\texttt{torch==2.1.2}) on a GPU in inference mode.
    \item \textbf{MLP classifier} implemented with Keras v. XXX using Keras Tune v XX
    \item \textbf{Containerization}: Docker image \texttt{tensorflow/tensorflow:2.14.0} was employed to ensure consistency and reproducibility for all jobs requiring GPU.
    \item \textbf{Vector Database}: Millvus (\texttt{milvus==2.3.0}) was used as the vector database to store embeddings and perform similarity search, interfaced through pymilvus (\texttt{pymilvus==2.3.6}).
\end{itemize}

\subsection{Appendix IV: The Practical Guide to Embedding Choice}
\label{sec:guide}

This section provides a practical framework for selecting and fine-tuning image embeddings based on our benchmarking results. By addressing common scenarios in e-Commerce and related domains, we offer actionable recommendations for balancing performance, computational efficiency, and task-specific requirements.

\subsubsection{Key Considerations for Embedding Selection}
When choosing an embedding model, it is essential to assess the following factors:

\begin{itemize}
    \item \textbf{Dataset Size and Diversity}: Large and diverse datasets benefit from full fine-tuning, whereas smaller datasets often perform well with pre-trained or top-tuned embeddings.
    \item \textbf{Computational Resources}: Resource-intensive models like ViT-L and ConvNeXt require significant training costs. Top-tuning or lighter architectures are more suitable for constrained environments.
    \item \textbf{Task Complexity}: Tasks involving fine-grained classification or retrieval with high inter-class similarity may necessitate full fine-tuning.
    \item \textbf{Label Availability}: When labeled data is scarce, self-supervised learning (SSL) embeddings and text-image embeddings with top-tuning provide a cost-efficient solution.
\end{itemize}

\subsubsection{Embedding Selection Based on Use Case}
Based on our benchmarking analysis, we recommend the following embedding strategies for common e-Commerce tasks:

\begin{table}[t]
    \centering
    \caption{Embedding strategies for common e-Commerce tasks.}

    \small
    \begin{tabular}{|p{2.8cm}|p{4cm}|p{5.5cm}|}
        \hline
        \textbf{Use Case} & \textbf{Recommended Strategy} & \textbf{Rationale} \\
        \hline
        Visual Search and Retrieval & Top-tuned text-image embeddings (e.g., CLIP, SigLip) & Achieves high retrieval accuracy while requiring minimal fine-tuning. Performs well across diverse product categories. \\
        \hline
        Product Categorization & Fully fine-tuned supervised embeddings (e.g., ViT, ConvNeXt) & Provides stable and high accuracy for structured classification tasks, outperforming SSL embeddings in category-based classification. \\
        \hline
        Cross-Domain Adaptation & Cross-tuning or top-tuning of text-image embeddings & Enables effective adaptation from one dataset to another. Works best when domains share visual characteristics. \\
        \hline
        Rapid Prototyping & Pre-trained text-image embeddings (e.g., CLIP, Marqo-B) & Strong zero-shot capabilities, requiring minimal adaptation for fast deployment. Useful for retrieval and tagging tasks. \\
        \hline
    \end{tabular}
    \label{tab:embedding_guide}
\end{table}

\subsubsection{Step-by-Step Guidelines}
To effectively implement the recommended strategies, follow these structured steps:

\subsubsection{Step 1: Define Task Requirements}
Clearly establish the primary objectives (e.g., classification, retrieval) and constraints such as available labeled data, computational budget, and deployment requirements.

\subsubsection{Step 2: Select the Embedding}
\begin{itemize}
    \item Choose \textbf{supervised embeddings} (e.g., ViT, ConvNeXt) for tasks requiring high precision and stability.
    \item Opt for \textbf{self-supervised embeddings} (e.g., DINOv2, MAWS) when adaptability across datasets is a priority.
    \item Consider \textbf{text-image contrastive models} (e.g., CLIP, SigLip, Marqo-B) for multimodal tasks and efficient retrieval.
\end{itemize}

\subsubsection{Step 3: Decide on Fine-Tuning Strategy}
\begin{itemize}
    \item Use \textbf{Full Fine-Tuning} when task-specific adaptation is crucial and computational resources allow.
    \item Choose \textbf{Top-Tuning} for an efficient trade-off between performance and cost, particularly for SSL and text-image embeddings.
    \item Explore \textbf{Cross-Tuning} when domain-specific labeled data is sparse but related datasets are available.
\end{itemize}

\subsubsection{Step 4: Train and Evaluate}
\begin{itemize}
    \item Train the embeddings using the selected fine-tuning strategy.
    \item Assess performance with relevant metrics (e.g., accuracy, MAP, Recall@1).
    \item Iterate by refining hyperparameters and model configurations as needed.
\end{itemize}

\subsubsection{Step 5: Deploy and Monitor}
Deploy the final model in production and continuously monitor its performance. Periodic fine-tuning or retraining may be necessary as new data is collected or task requirements evolve.

\subsubsection{Trade-Offs and Recommendations}
Table~\ref{tab:trade_offs} outlines the trade-offs between fine-tuning strategies, helping to balance performance, efficiency, and domain adaptability.

\begin{table}[t]
    \centering
    \caption{Trade-offs between fine-tuning strategies.}
    \small
    \begin{tabular}{|p{3cm}|p{5cm}|p{5cm}|}
        \hline
        \textbf{Strategy} & \textbf{Advantages} & \textbf{Limitations} \\
        \hline
        Full Fine-Tuning & Highest performance, strong task adaptation & Computationally expensive, risk of overfitting. \\
        \hline
        Top-Tuning & Cost-efficient, significantly improves SSL and text-image models & May not fully match fine-tuned supervised embeddings for classification. \\
        \hline
        Cross-Tuning & Enables knowledge transfer across domains, useful for text-image models & Performance varies depending on dataset similarity; top-tuning is often more effective. \\
        \hline
    \end{tabular}
   
    \label{tab:trade_offs}
\end{table}

\subsubsection{Conclusion}
This guide provides a structured approach to embedding selection and fine-tuning for e-Commerce applications \cref{tab:embedding_guide}. Our findings highlight the growing role of text-image embeddings in retrieval and classification, often outperforming supervised models with minimal adaptation. While full fine-tuning remains the strongest approach for domain-specific classification, top-tuning of SSL and contrastive models offers an efficient alternative. Future work should explore hybrid fine-tuning strategies that dynamically adjust between full fine-tuning and top-tuning based on dataset characteristics.

\subsection{Appendix V: Limitations}

While our study provides a comprehensive analysis, several limitations should be acknowledged:

\subsubsection{Scope of Datasets and Models}
Although our work spans six diverse e-Commerce datasets, there may be unique challenges in other domains that were not captured. Similarly, while we evaluated a broad range of models, certain state-of-the-art architectures (e.g., multi-modal models) were not included.

\subsubsection{Generalizability of Cross-Tuning}
The effectiveness of cross-tuning depends heavily on the similarity between source and target datasets. Our findings, while indicative, may not generalize to all cross-domain scenarios. Further exploration of this strategy in diverse domains is necessary.

\subsubsection{Computational Constraints}
The resource-intensive nature of some fine-tuning experiments limited the depth of our analysis in certain configurations, particularly for large-scale datasets. Future work could explore additional optimizations to address these constraints.

\end{document}